\documentclass[10pt,twocolumn,letterpaper]{article}

\usepackage{cvpr}
\usepackage{times}
\usepackage{epsfig}
\usepackage{epstopdf}
\usepackage{graphicx}
\usepackage{amsmath}
\usepackage{mathtools}
\usepackage{amssymb}
\usepackage{booktabs}
\usepackage{makecell}
\usepackage{multirow}
\usepackage{cite}
\newcommand{\PreserveBackslash}[1]{\let\temp=\\#1\let\\=\temp}
\newcolumntype{C}[1]{>{\PreserveBackslash\centering}p{#1}}
\newcolumntype{R}[1]{>{\PreserveBackslash\raggedleft}p{#1}}
\newcolumntype{L}[1]{>{\PreserveBackslash\raggedright}p{#1}}
\DeclareMathOperator*{\argmax}{argmax}

\cvprfinalcopy 

\def\cvprPaperID{12} 

\ifcvprfinal\fi
\pdfoutput=1
\begin{document}

\title{SANet: Structure-Aware Network for Visual Tracking}

\author{Heng Fan \;\;\;\; Haibin Ling\\
$^{1}$Department of Computer and Information Sciences, Temple University, Philadelphia, USA\\
$^{2}$HiScene Information Technologies, Shanghai, China\\
{\tt\small \{hengfan, hbling\}@temple.edu}
}

\maketitle

\begin{abstract}
Convolutional neural network (CNN) has drawn increasing interest in visual tracking owing to its powerfulness in feature extraction. Most existing CNN-based trackers treat tracking as a classification problem. However, these trackers are sensitive to similar distractors because their CNN models mainly focus on inter-class classification. To address this problem, we use self-structure information of object to distinguish it from distractors. Specifically, we utilize recurrent neural network (RNN) to model object structure, and incorporate it into CNN to improve its robustness to similar distractors. Considering that convolutional layers in different levels characterize the object from different perspectives, we use multiple RNNs to model object structure in different levels respectively. Extensive experiments on three benchmarks, OTB100, TC-128 and VOT2015, show that the proposed algorithm outperforms other methods. Code is released at www.dabi.temple.edu/$\sim$hbling/code/SANet/SANet.html.
\end{abstract}

\section{Introduction}
\label{secIntrod}

Object tracking is one of the most important components in computer vision and has a variety of applications such as video surveillance, robotics, human-computer interaction and so forth \cite{yilmaz2006object}. Despite great progress in recent decades, visual tracking remains a challenging task due to appearance changes caused by deformation, illumination variations, occlusion and so on. 

The deep neural networks~\cite{lecun1989backpropagation}, which demonstrate the powerfulness in extracting high-level feature representations~\cite{girshick2014rich}, have drawn extensive attention in computer vision, such as image classification~\cite{krizhevsky2012imagenet}, recognition~\cite{simonyan2014very}, saliency detection~\cite{wang2016saliency}, semantic segmentation~\cite{long2015fully} and so on. Inspired by this, many CNN-based trackers \cite{fan2010human,nam2015learning,ma2015hierarchical,hong2015online,li2016deeptrack,wang2015visual,danelljan2016beyond,tao2016siamese} have been proposed. Among them, \cite{nam2015learning} presents an on-line tracking method based on a multi-domain CNN architecture and achieves state-of-the-art performances on various benchmarks. By leveraging extensive annotated videos, it learns a robust shared representation to classify object from background. However, this tracker may be sensitive to similar distractors because the learned CNN model mainly focuses on inter-class classification. In the presence of distractors, the tracker has a high chance to misclassify the object and background.

\begin{figure*}[!t]
\centering
\includegraphics[width=\linewidth,height=.21\linewidth]{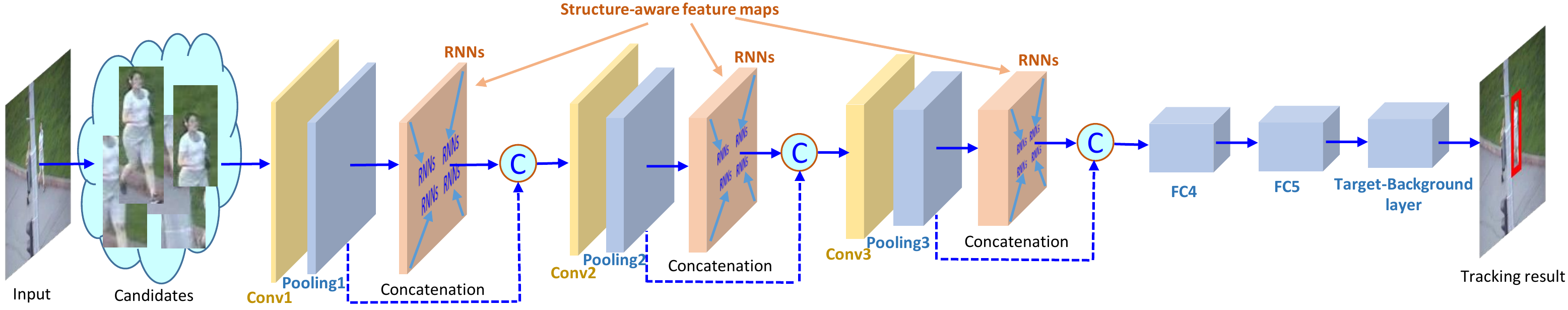}\\
\caption{Illustration of the proposed SANet for visual tracking.}\label{fig:SANet.archi}
\end{figure*}

Recently, recurrent neural networks (RNNs) \cite{elman1990finding}, which show great success in neural language process (NLP) \cite{graves2012sequence}, have been brought to the computer vision community \cite{shuai2015dag,zuo2016learning,zuo2017combining,fan2016multi,byeon2015scene,van2016pixel,cui2016recurrently} owing to the capability of capturing
long-range dependencies among sequential data. With this property, RNNs are able to model the self-structure of object.

Inspired by the above observations, in this paper we propose a novel Structure-Aware Network (SANet) architecture for visual tracking by utilizing RNNs to model self-structure of object. Different from conventional CNNs in tracking, which mainly pay attention to inter-class classification and thus are prone to drift in presence of similar distractors, our SANet leverages RNNs to encode self-structure of object during learning, which helps improve our model in discriminating not only background objects of inter-class but also similar distractors of intra-class. Because when similar distractors occur, our model is able to capture even slight difference between the reference and distractors, and use the discrepancies to distinguish object from distractors. Taking into account that convolutional layers at different levels characterize the object from different perspectives, we apply multiple RNNs to modeling structure of object in different levels respectively, which strengthens robustness of the proposed model. Besides, to supply our SANet with richer information, we adopt a skip concatenation strategy to fuse CNN and RNN feature maps, and demonstrate its effectiveness in improving performance. Figure \ref{fig:SANet.archi} illustrates the proposed method in this paper. Extensive experimental results on two large-scale tracking benchmarks demonstrate the advances of our method.

In summary, we make the following contributions:

\begin{itemize}
\item We propose the structure-aware network architecture for tracking by using RNNs to encode self-structure of object during learning, which helps our model improve not only the capability of discriminating background objects of inter-class but also similar distractors of intra-class.
\item To supply our networks with richer information, we adopt a skip concatenation strategy to fuse CNN and RNN features, and show its effectiveness in improving tracking performance.
\item Extensive experiments on three large-scale tracking benchmarks, OTB100 \cite{wu2015object}, TC-128 \cite{liang2015encoding} and VOT2015 \cite{kristan2015visual}, demonstrate that the proposed tracker outperforms other state-of-the-art methods.
\end{itemize}

The rest of this paper is organized as follows. Section \ref{secRel} briefly summarizes related work. Section~\ref{secRNN} illustrates self-structure modeling of object with RNNs. Section~\ref{secTracking} introduces the proposed tracking algorithm in details. Experiments are described in Section~\ref{secExp}, followed by conclusion in Section~\ref{secCon}.

\section{Related Work}
\label{secRel}
Object tracking is one of the most challenging problems in computer vision and has been extensively studied \cite{yilmaz2006object}. In the following we highlight three lines of works which are most related to ours.

\vspace{-2mm}\paragraph{\bf Visual tracking:} Roughly speaking, tracking algorithms can be categorized into two types: discriminative methods \cite{zhang2014fast,henriques2015high,babenko2009visual,wang2011superpixel,danelljan2015learning,kalal2012tracking} and generative methods \cite{mei2011robust,fan2016robust,fan2015robust,bao2012real,ross2008incremental,wang2014visual,zhang2016defense,kwon2010visual}. Discriminative methods regard tracking as a classification problem which aims to separate object from ever-changing background. These methods employ both the foreground and background information to learn classifiers via P-N learning \cite{kalal2012tracking}, multiple instance learning (MIL) \cite{babenko2009visual}, correlation filters \cite{danelljan2015learning,henriques2015high} and so forth. On the contrary, generative approaches formulate the
tracking problem as searching for regions most similar to the target object. These methods are based on either subspace models or templates and update appearance model dynamically. Some representative generative methods includes incremental subspace learning \cite{ross2008incremental}, sparse representation \cite{mei2011robust,bao2012real,zhang2016defense}, probabilistic model \cite{wang2014visual,kwon2010visual} and so on.

Despite promising results for tracking in some constrained situations, the performances of aforementioned approaches are vulnerable due to the limitation of low-level hand-crafted features in complex environments where object appearances are simultaneously affected by various factors (e.g., motion blur, occlusion, deformation, scale changes, illumination variations). One possible solution is to adopt the learned high-level features for object appearance representation.

\begin{figure*}[!htbp]
\centering
\includegraphics[width=\linewidth,height=.13\linewidth]{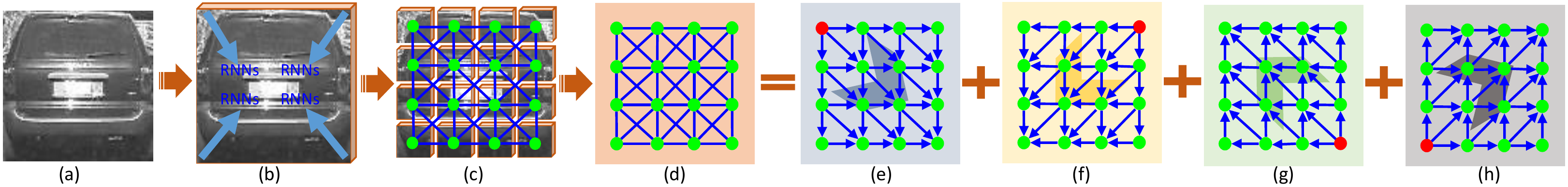}\\
\caption{Decomposition of undirected cyclic graph into four directed acyclic graphs. Images (a) and (b) are inputs. Self-structure of object is encoded in an undirected cyclic graph in images (c) and (d). Images (e), (f), (g) and (h) are four directed acyclic graphs along southeast, southwest, northwest and northeast directions.}\label{fig:RNN-decomposition}
\end{figure*}

\vspace{-2mm}\paragraph{\bf Deep networks in tracking:} Owing to the powerfulness in feature extraction, deep networks haven been introduced into visual tracking. \cite{fan2010human} proposes a human-tracking method based on CNNs. \cite{wang2013learning} introduces a deep compact tracker based on stacked autoencoder. \cite{li2016deeptrack} presents an on-line learning method based on a pool of CNNs. However, these trackers suffer from lack of enough training data to learn a robust representation, which degrades the performance of tracker. To address this problem, \cite{ma2015hierarchical,hong2015online,wang2015visual,danelljan2016beyond} transfer CNNs pretrained on a large-scale dataset for image classification, however, the representation may not be very effective due to the fundamental difference between classification and tracking tasks \cite{nam2015learning}. To deal with this issue, \cite{nam2015learning,tao2016siamese} propose to train the CNNs on a set of annotated video sequences, and showed that the CNNs trained on video sequences are more robust. In particular, \cite{nam2015learning} introduces an effective strategy, i.e., multi-domain learning \cite{duan2009domain}, to train the CNNs, which helps to discriminate object from background. However, this method is sensitive to similar distractors because its CNN model mainly concentrate on inter-class classification. Different from \cite{nam2015learning}, we use RNNs to model self-structure of object and encode it into CNN, which is beneficial to distinguish distractors of intra-class.

\vspace{-2mm}\paragraph{\bf RNNs on image processing:} RNNs \cite{elman1990finding} have been first introduced to handle sequential prediction task \cite{graves2012sequence}, and then extended to multi-dimensional image processing tasks \cite{GravesFS07} such as image classification \cite{zuo2016learning}, scene labeling \cite{shuai2015dag,byeon2015scene}, person re-identification \cite{varior2016siamese} and so on. By capturing long-range dependencies among image units, RNNs are able to well model self-structure of object.

For visual tracking, one major challenge is to separate object from similar distractors of intra-class. CNNs cannot well deal with this situation because CNNs mainly focus on classifying objects belonging to different classes. One possible solution is to leverage the differences between intra-class objects to separate them. To make CNNs aware of the difference between objects of intra-class, we utilize RNNs to model self-structure of object and encode it into CNNs for classification. With this structural information, it is able to discriminate object from similar distractors via their even slight discrepancies.

We note that RNNs have been investigated in \cite{cui2016recurrently} for tracking, but it is different from ours. In \cite{cui2016recurrently}, RNNs are used to model spatial-relationship between object and surrounding background, and obtain a confidence map to regularize correlation filters. However, in our work, we apply RNNs to modeling structure of object itself, and use such structure information to discriminate distractors of intra-class. Besides, the RNNs in this work are integrated with CNNs, and trained with enough video sequences. While in \cite{cui2016recurrently}, RNNs are only trained with a few initial frames, and updated with each frame, which may not fully explore the advantages of RNNs.

\section{RNNs for Object Self-Structure Modeling}
\label{secRNN}
RNNs~\cite{elman1990finding} are developed for modeling dependencies in sequential data. Given an input sequence $\{ x^{(t)} \}_{t=1,2,\cdots,T}$ of length $T$, the hidden layer $h^{(t)}$ and output layer $y^{(t)}$ at each time step $t$ are calculated with
\begin{equation}\label{eq1}
\begin{cases}
h^{(t)}=\phi(Ux^{(t)}+Wh^{(t-1)}+b)  \\
y^{(t)}=\sigma (Vh_{t}+c)
\end{cases}
\end{equation}
where $U$, $W$ and $V$ represent weight matrices between the input and the hidden layer, the previous hidden layer and the current hidden layer, and the hidden layer and the output layer respectively; $b$ and $c$ represent bias terms; and $\phi(\cdot)$ and $\sigma(\cdot)$ are non-linear activation functions. Since the inputs are progressively stored in hidden layers, RNNs can model long-range contextual dependencies among the sequence elements.

For two-dimensional image data, different from one-dimension sequential data, its self-structure is encoded in an undirected cyclic graph (see Figure \ref{fig:RNN-decomposition}(c)). Because of the loopy structure of undirected cyclic graph, the aforementioned RNNs cannot be directly applied to images. To handle this issue, we approximate the topology of an undirected cyclic graph by the combination of several directed acyclic graphs as in \cite{shuai2015dag}, and use variant RNNs to model self-structure of the target object as shown in Figure~\ref{fig:RNN-decomposition}.

Assume that a directed acyclic graph is represented with $\mathcal{G}=\{ {\mathcal{V}, \mathcal{E}} \}$, where $\mathcal{V}=\{ v_{i} \}_{i=1,2,\cdots,N}$ denotes vertex set and $\mathcal{E}=\{ e_{ij} \}$ is the edge set, in which $e_{ij}$ represents a directed edge from $v_{i}$ to $v_{j}$. The structure of RNNs follows the same topology as $\mathcal{G}$. A forward propagation sequence can be seen as traversing $\mathcal{G}$ from the start point, and each vertex relies on its all predecessors. For vertex $v_{i}$, therefore, the hidden layer $h^{(v_{i})}$ is expressed as a non-linear function over current input $x^{(v_{i})}$ at $v_{i}$ and summation of hidden layers of all its predecessors. Specifically, the hidden layer $h^{(v_{i})}$ and output layer $y^{(v_{i})}$ at each $v_{i}$ are computed with
\begin{equation}\label{eq2}
\begin{cases}
h^{(v_{i})} = \phi (Ux^{(v_{i})}+W~\sum\limits_{\mathclap{{v_{j}\in{\mathcal{P}_{\mathcal{G}}(v_{i})}}}}{h^{(v_{j})}}+b) \\
y^{(v_{i})} = \sigma (Vh^{(v_{i})}+c)
\end{cases}
\end{equation}
where $\mathcal{P}_{\mathcal{G}}(v_{i})$ denotes the predecessor set of $v_{i}$ in $\mathcal{G}$.

The forward pass of RNNs can be calculated with Eq. (\ref{eq2}). For backward propagation, we need to calculate derivatives at each vertex in the RNNs. For each vertex in the directed acyclic graph, it is processed in the reverse order of forward propagation sequence. In details, to compute the derivatives at vertex $v_{i}$, we need to look at the forward passes of all its successors. Let $\mathcal{S_{G}}(v_{i})$ denote the direct successor set for $v_{i}$ in $\mathcal{G}$. For each $v_{k}\in {\mathcal{S_{G}}(v_{i})}$, its hidden layer is computed by
\begin{equation}\label{eq3}
\begin{cases}
h^{(v_{k})} = \phi (Ux^{(v_{k})}+Wh^{(v_{i})}+\sum\limits_{\mathclap{v_{l}\in{\mathcal{Q}}}}{Wh^{(v_{l})}}+b) \\
y^{(v_{k})} = \sigma (Vh^{(v_{k})}+c)
\end{cases}
\end{equation}
where $\mathcal{Q}=\mathcal{P}_{\mathcal{G}}(v_{k})-\{ {v_{i}}\}$. Combining Eq (\ref{eq2}) and (\ref{eq3}), we can see that the errors back-propagated to the hidden layer at $v_{i}$ come from two sources: directed errors from $v_{i}$ (i.e., $\frac{\partial y^{(v_{i})}}{\partial h^{(v_{i})}}$) and summation over indirected errors from all its successors
$v_{k} \in \mathcal{S_{G}}(v_{i})$ (i.e., $\sum\nolimits_{v_{k}}{\frac{\partial y^{(v_{k})}}{\partial h^{(v_{i})}}}=\sum\nolimits_{v_{k}}{\frac{\partial y^{(v_{k})}}{\partial h^{(v_{k})}}}{\frac{\partial h^{(v_{k})}}{\partial h^{(v_{i})}}}$). Therefore, the derivatives at vertex $v_{i}$ can be obtained by
\begin{equation}\label{eq4}
\begin{cases}
\mathrm{d}h^{(v_{i})}&=V^{T}\sigma{'}(y^{(v_{i})})+~\sum\limits_{\mathclap{v_{k}\in{\mathcal{S_{G}}(v_{i})}}}{W^{T}\mathrm{d}h^{(v_{k})}\circ{\phi{'}(h^{(v_{k})})}}\\
\nabla W^{(v_{i})}&=~~\sum\limits_{\mathclap{v_{k}\in{\mathcal{S_{G}}(v_{i})}}}~\mathrm{d}h^{(v_{k})}\circ{\phi{'}(h^{(v_{k})})}(h^{(v_{i})})^{T}\\
\nabla U^{(v_{i})}&=\mathrm{d}h^{(v_{i})}\circ{\phi{'}(h^{(v_{i})})}(x^{(v_{i})})^{T}\\
\nabla b^{(v_{i})}&=\mathrm{d}h^{(v_{i})}\circ{\phi{'}(h^{(v_{i})})}\\
\nabla V^{(v_{i})}&=\sigma{'}(y^{(v_{i})})(h^{(v_{i})})^{T} \\
\nabla c^{(v_{i})}&=\sigma{'}(y^{(v_{i})})
\end{cases}
\end{equation}
where $\circ$ is the Hadamard product, $\sigma{'}(\cdot)=\frac{\partial L}{\partial y(\cdot)}\frac{\partial y(\cdot)}{\partial \sigma}$ is the derivative of loss function $L$ with respect to output function $\sigma$, and $\phi{'}(\cdot)=\frac{\partial h}{\partial \phi}$. Note that the superscript $T$ denotes transposition operation.

With Eq (\ref{eq2}) and (\ref{eq4}), we can perform forward and backward passes on one directed acyclic graph. In this paper, we decompose the undirected cyclic graph into four directed acyclic graphs along southeast, southwest, northwest and northeast directions. Figure \ref{fig:RNN-decomposition} visualizes the decomposition. Let $\mathcal{G^{U}}=\{ \mathcal{G}_{1},\mathcal{G}_{2},\mathcal{G}_{3},\mathcal{G}_{4} \}$ denote the undirected cyclic graph, where $\mathcal{G}_{1},\mathcal{G}_{2},\mathcal{G}_{3},\mathcal{G}_{4}$ represent the four directed acyclic graphs respectively. For each $\mathcal{G}_{m}$ ($m=1,2,3,4$), we can get the corresponding hidden layer $h_{m}$ by performing RNNs. The summation of all hidden layers are fed to the output layer. We use Eq (\ref{eq5}) to express this process
\begin{equation}\label{eq5}
\begin{cases}
h_{m}^{(v_{i})} = & \phi (U_{m}x^{(v_{i})}+~~\sum\limits_{\mathclap{v_{j}\in{\mathcal{P}_{\mathcal{G}_{m}}(v_{i})}}}{W_{m}{h_{m}^{(v_{j})}}}  +b_{m}) \\
y^{(v_{i})} = &\sigma (~\sum\limits_{\mathclap{\mathcal{G}_{m}\in{\mathcal{G^{U}}}}}~{V_{m}h_{m}^{(v_{i})}}+c)
\end{cases}
\end{equation}
where $U_{m}$, $W_{m}$, $V_{m}$, and $b_{m}$ are matrix parameters and bias term for $\mathcal{G}_{m}$, $c$ is the bias term for final output, and $\mathcal{P}_{\mathcal{G}_{m}}(v_{i})$ denotes the predecessor set of $v_{i}$ in $\mathcal{G}_{m}$. The error back-propagated to previous convolutional layer at $v_{i}$ is computed by
\begin{equation}\label{eq6}
\nabla x^{(v_{i})} = \sum\limits_{\mathclap{\mathcal{G}_{m}\in{\mathcal{G^{U}}}}}U_{m}^{T}\mathrm{d}h_{m}^{(v_{i})}\circ{\phi{'}(h_{m}^{(v_{i})})}
\end{equation}

\section{Proposed Tracking Algorithm}
\label{secTracking}
\subsection{Network architecture}
The architecture of the proposed network is depicted in Figure~\ref{fig:SANet.archi}, which receives a 107$\times$107 (same in \cite{nam2015learning}) RGB input, and has three convolutional layers (each with ReLU and pooling layers), two fully connected layers and one fully connected classification layer. Each pooling layer is followed by a recurrent layer, which models the structure of object in this level. Besides, to provide the next convolutional layer with more information, we adopt a skip concatenation strategy to fuse the features from pooling and recurrent layers.

\begin{figure*}[!t]
\centering
\includegraphics[width=.5\linewidth,height=.31\linewidth]{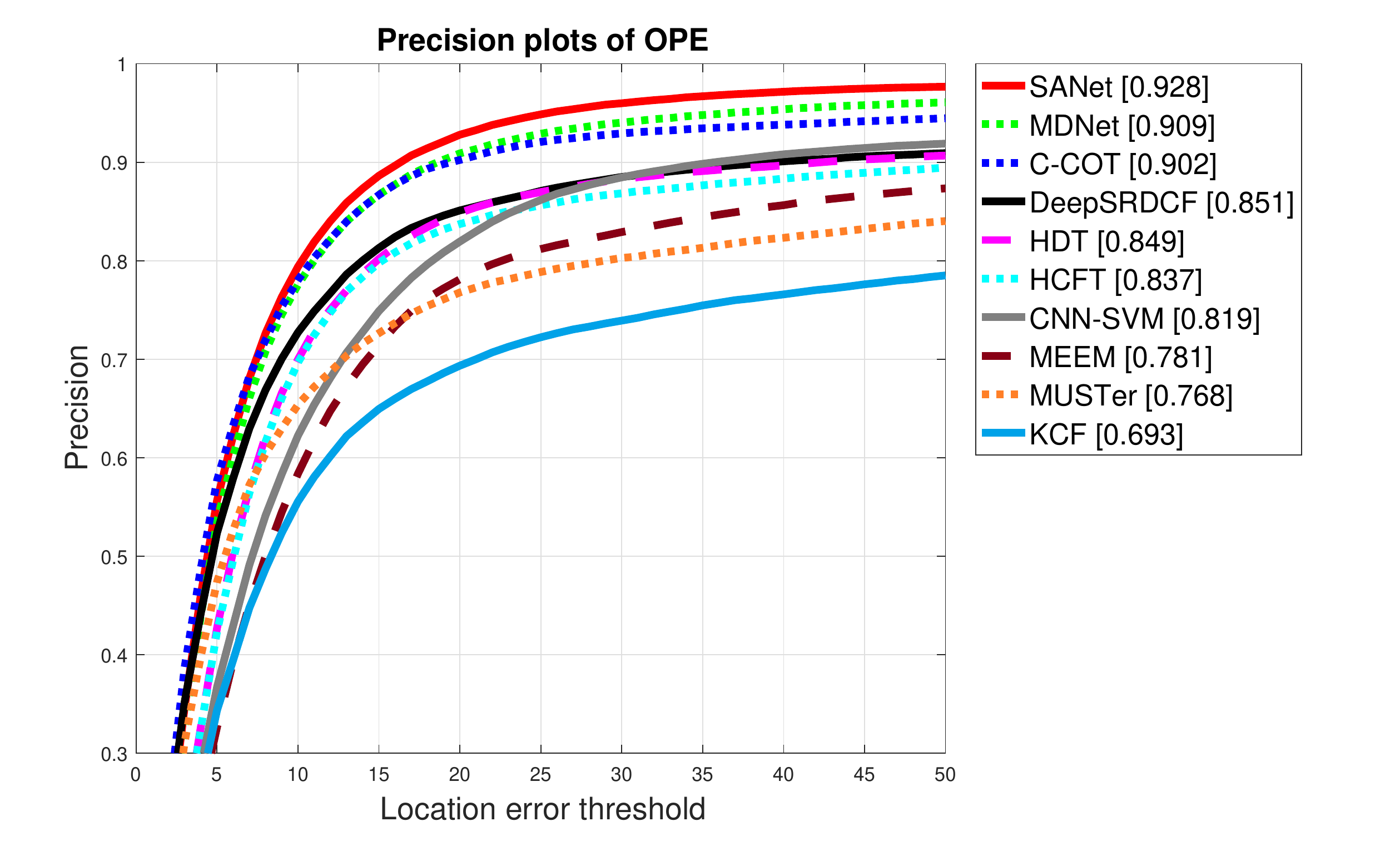}\includegraphics[width=.5\linewidth,height=.31\linewidth]{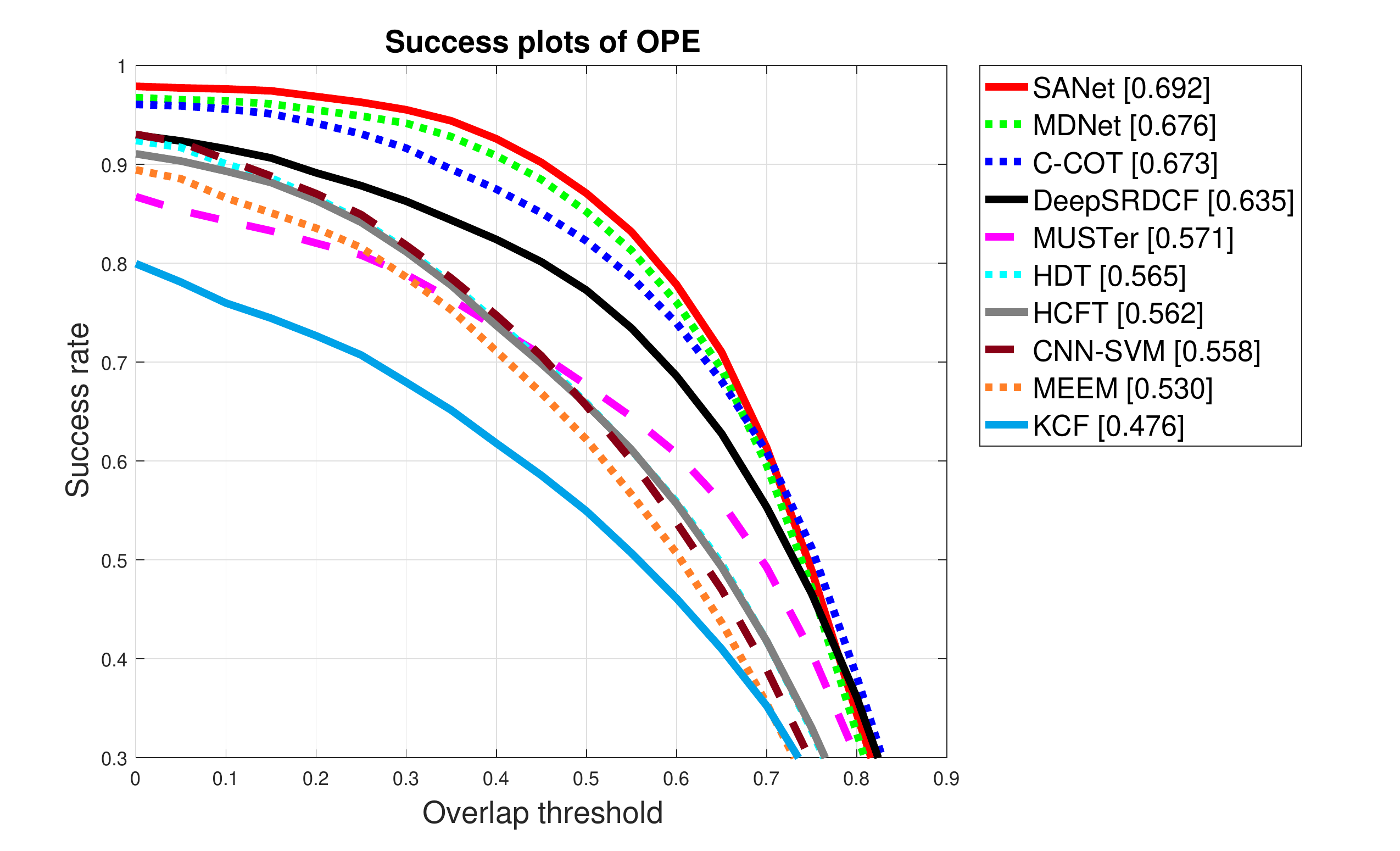}\\
\caption{Precision and success plots on OTB100 \cite{wu2015object}. The numbers in the legend indicate the representative precisions at 20 pixels for precision plots, and the area-under-curve scores for success plots.}\label{fig4}
\end{figure*}

\begin{figure*}[!t]
\centering
\includegraphics[width=\linewidth,height=.245\linewidth]{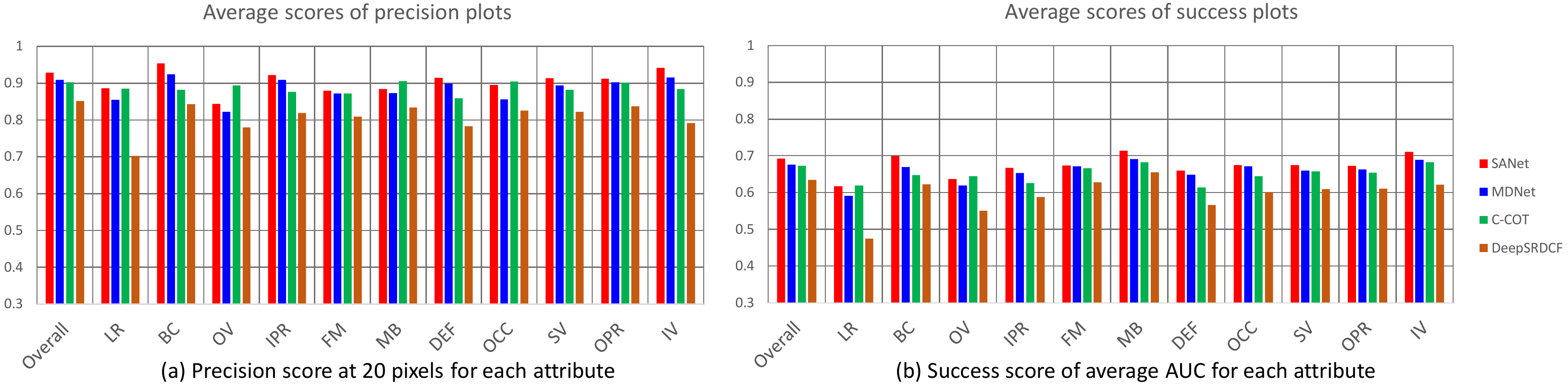}\\
\caption{Precision scores at 20 pixels and success scores of average AUC of the four leading trackers under different attributes of test sequences in OPE on OTB \cite{wu2015object}, including illumination variation (IV), out-of-plane rotation (OPR), scale variation (SV), occlusion (OCC), deformation (DEF), motion blur (MB), fast motion (FM), in-plane rotation (IPR), out-of-view (OV), background cluttered (BC) and low resolution (LR).}\label{fig5}
\end{figure*}

\subsection{Training}
Inspired by the success in \cite{nam2015learning}, we utilize a set of annotated video sequences to train the whole network. For convolutional layer, it is trained by the Stochastic Gradient Descent (SGD) method, and the recurrent layer is trained by the method introduced in Section \ref{secRNN}. Besides, we also adopt the multi-domain learning strategy as in \cite{nam2015learning}. In the training stage, the final layer has $K$ branches, and only the $k^{th}$ branch is handled in the $k^{th}$ iteration. The whole training process ends when the network converges or a predefined max number of iteration is reached. In the testing stage, the $K$ branches of the final layer are replaced with a single branch corresponding to the tracked object. By adopting the multi-domain strategy, the performance of the proposed tracker is further improved.

\subsection{Tracking and update}
Visual tracking is achieved within the particle filter framework. For each new frame, we sample $N$ target candidates $\{c_{i}\}_{i=1}^{N}$ around the position of target in last frame, and evaluate them by their positive scores $p(c_{i})$ obtained by the network. The positive score of each candidate indicates its probability belonging to target class. The candidate with the highest positive score is chosen to be the tracked result $O$ as follows
\begin{equation}\label{eq7}
O=\mathop{\argmax}_{c_{i}}\;p(c_{i})
\end{equation}

Due to object appearance variation caused by factors such as lighting change and deformation, update is essential during tracking. We adopt two strategies to update the network as in \cite{nam2015learning}: short-term and long-term updates. When the positive score $p(O)$ of the tracked result is smaller than a predefined threshold $\theta$, the short-term update is performed. Otherwise, the long-term update is executed. For the long-term update, the whole network is updated with the collected positive samples for a long period of time and negative samples stored for a short period time. While for the short-term update, both positive and negative samples for update are collected from a short period of time.

\subsection{Hard minibatch mining}
In tracking, most negative samples are redundant, and only a few distracting negative samples are helpful in training a discriminative classifier. In this situation, the plain SGD method easily results in drift due insufficient effective negative samples. To address this problem, \cite{nam2015learning} leverages a popular solution, i.e., hard negative mining, in object detection \cite{sung1998example}. In this paper, we utilize the same strategy to alleviate this problem.

\subsection{Box refinement}

To locate the target object, we sample multiple positive samples around the target, which may result in failure to find the tight boxes enclosing the target. To handle this issue, \cite{nam2015learning,tao2016siamese} adopt a refinement step in each frame to improve the predicted bounding box. In this paper, the same strategy is utilized. In the first frame, we train a simple linear regression model to predict the position of target. In subsequent frames, we use the regression model to adjust the target locations obtained by Eq. (\ref{eq3}) if the positive score of the tracked result is larger than $\theta$.

\section{Experiments}
\label{secExp}

\subsection{Implementation details}
The proposed method is implemented in Matlab based on MatConvNet \cite{vedaldi15matconvnet}, and runs at around 1 frames per second (FPS) with 3.7 GHz Intel i7 Core and a NVIDIA GTX TITAN Z GPU. In each new frame, we sample 300 ($N=300$) target candidates in translation and scale dimension from a Gaussian distribution. Three independent RNNs are utilized to model image unit dependencies in multiple levels, i.e., the 1$^{st}$, 2$^{nd}$ and 3$^{rd}$ pooling layers. The dimension of hidden layers of RNNs are set to the same as the channels of the 1$^{st}$, 2$^{nd}$ and 3$^{rd}$ pooling layers. The learning rates of RNNs are initialized to be 10$^{-3}$ and decay exponentially with the rate of 0.9. Other parameters of convolutional layers are set to the same as in \cite{nam2015learning}.

\subsection{Evaluation on OTB}
\begin{figure*}[!htbp]
\centering
\includegraphics[width=.495\linewidth,height=.31\linewidth]{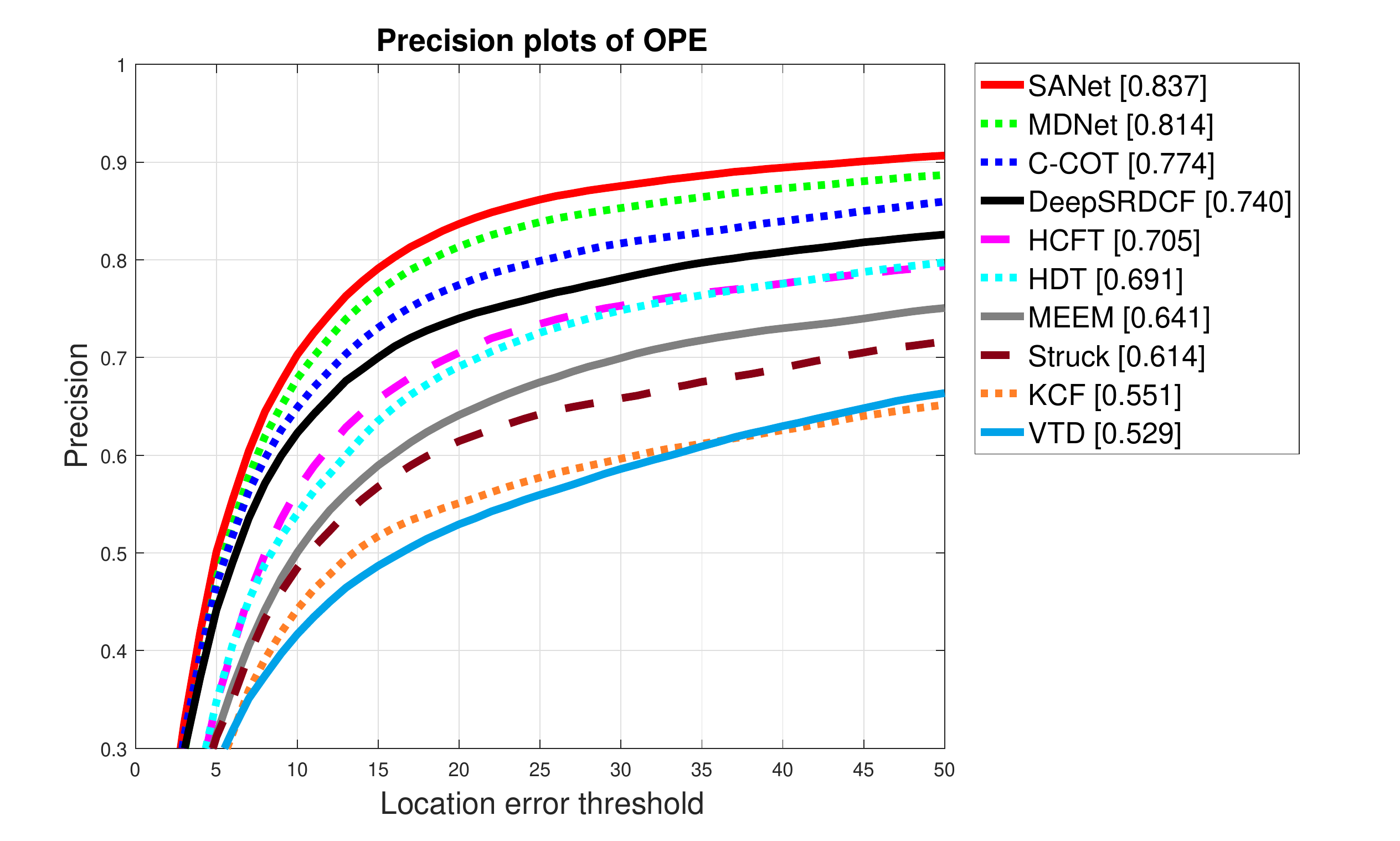}\hfill\includegraphics[width=.495\linewidth,height=.31\linewidth]{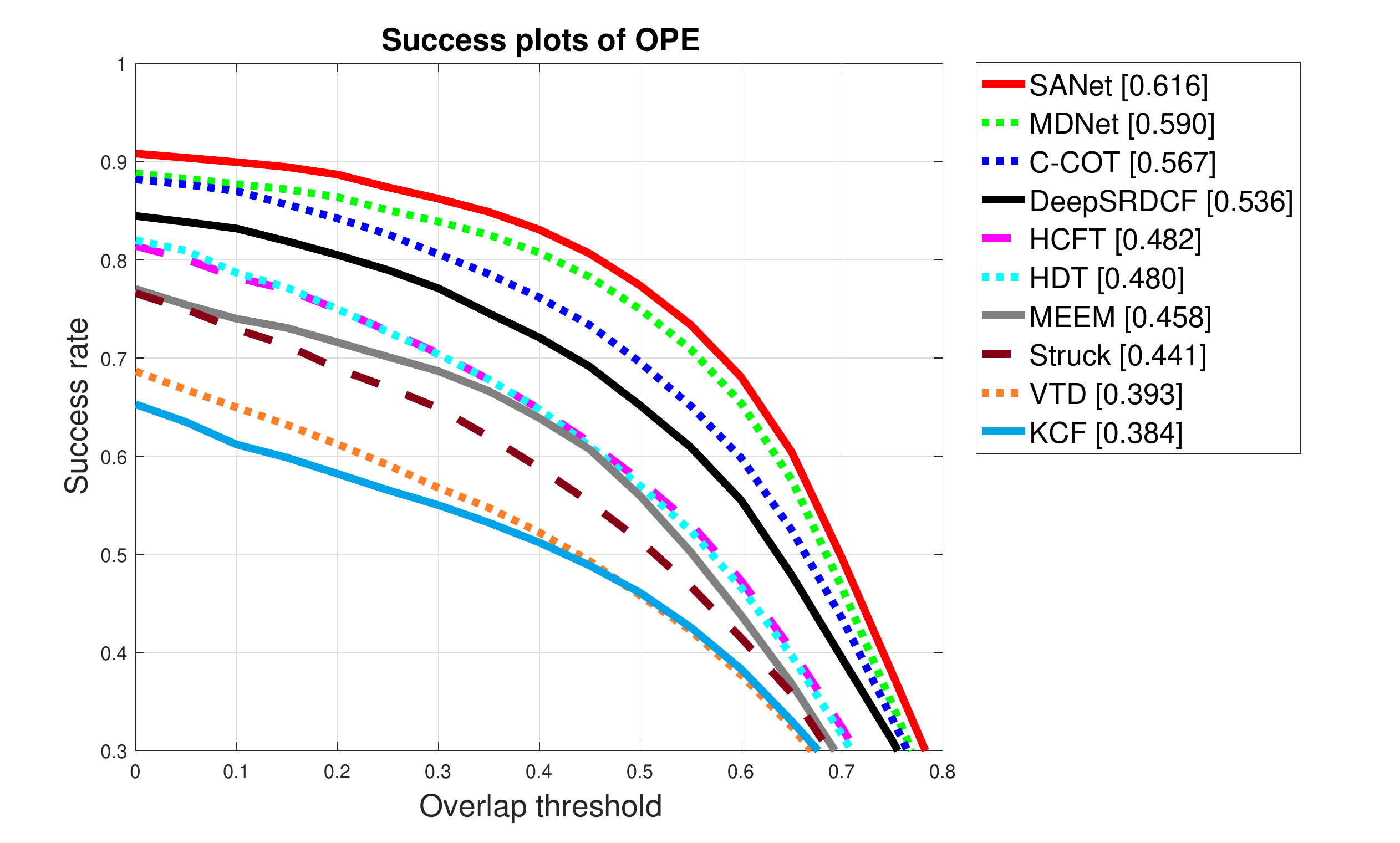}\\
\caption{Precision and success plots on TC-128 \cite{liang2015encoding}. The numbers in the legend indicate the representative precisions at 20 pixels for precision plots, and the area-under-curve scores for success plots.}\label{fig8}
\end{figure*}
\begin{figure*}[!t]
\centering
\includegraphics[width=\linewidth]{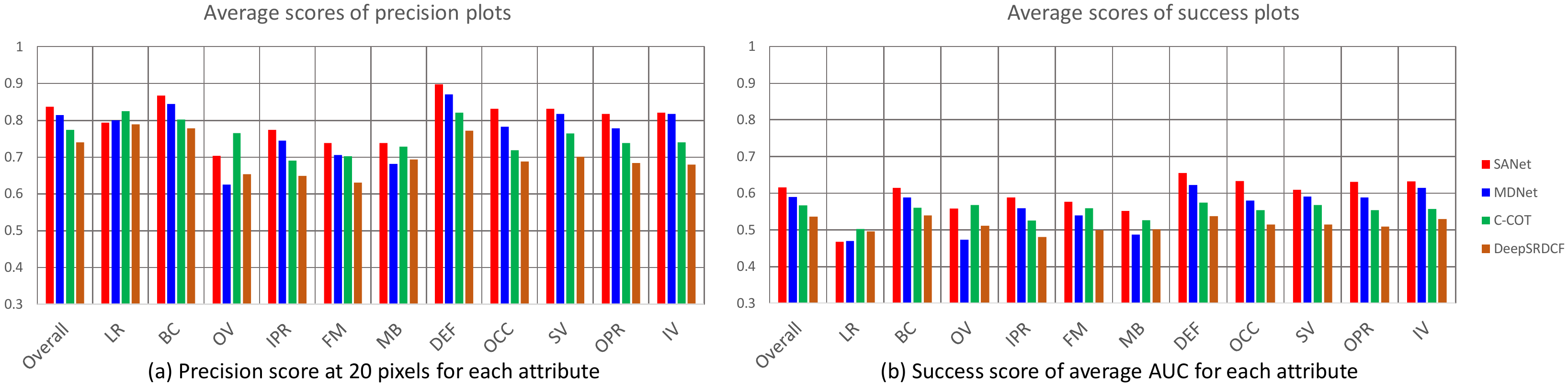}\\
\caption{Precision scores at 20 pixels and success scores of average AUC of the four leading trackers under different attributes of test sequences in OPE on TC-128 \cite{liang2015encoding}, including: IV, OPR, SV, OCC, DEF, MB, FM, IPR, OV, BC and LR.}\label{fig9}
\end{figure*}

OTB100~\cite{wu2015object} is a popular tracking benchmark containing 100 fully annotated videos with various challenges. We employ the precision plots and success plots defined in \cite{wu2015object} to evaluate the robustness of the tracking approaches. In addition to the trackers included in the benchmark \cite{wu2015object}, e.g., SCM \cite{zhong2012robust} and Struck \cite{hare2011struck}, we also compare our method with most recent state-of-the-art trackers including MEEM \cite{zhang2014meem}, TGPR \cite{gao2014transfer}, MDNet \cite{nam2015learning}, MUSTer \cite{hong2015multi}, CNN-SVM \cite{hong2015online}, DeepSRDCF \cite{danelljan2015learning}, C-COT \cite{danelljan2016beyond}, HCFT \cite{ma2015hierarchical}, HDT \cite{qi2016hedged} and KCF \cite{henriques2015high}. To train the network, we utilize image sequences collected from VOT2013 \cite{kristan2013visual}, VOT2014 \cite{kristan2014visual} and VOT2015 \cite{kristan2015visual}, excluding the videos included in OTB \cite{wu2015object}.

Figure \ref{fig4} shows the comparisons of our method with other state-of-the-art tracker in terms of precision and success plots, respectively. From Figure \ref{fig4}, we can see that the proposed approach outperforms other state-of-the-art trackers in both measures. The exceptional scores at mild thresholds means our tracker hardly misses targets while the competitive scores at strict thresholds implies that our algorithm also finds tight bounding boxes to targets. Among other trackers, \cite{nam2015learning} also utilizes deep convolutional neural networks to learn the object appearance representation. However, it does not take self-structure information of object into account. While our method considers structure of object during learning, and improves the  ability of network to distinguish object from background. Figure \ref{fig5} illustrates that our tracker is able to effectively deal with various challenging situations. It is worth noticing that, compared with the method in \cite{nam2015learning}, our method improves performance of tracking in all 11 attributes.


In \cite{nam2015learning}, an effective strategy, i.e., multi-domain learning, is adopted to train the networks. In this work, we also leverage this strategy to train the networks. To verify the impact of multi-domain learning, we conduct another experiments without multi-domain learning method to train the network, while keep other conditions the same. Without multi-domain learning, our method achieves 0.922 ranking score in precision plots and 0.688 ranking score in success plots. Compared with using multi-domain learning method, the tracking performance slightly degrades, which demonstrates the effectiveness of multi-domain learning strategy.

\subsection{Evaluation on TC-128}

TC-128~\cite{liang2015encoding} contains 128 fully annotated color image sequences. We use the same metrics used in \cite{wu2015object} and \cite{liang2015encoding}, i.e., precision and success plots, to evaluate the tracking methods. In addition to the trackers tested in the benchmark \cite{liang2015encoding}, we add some recent trackers including \cite{danelljan2015learning}, C-COT \cite{danelljan2016beyond}, HCFT \cite{ma2015hierarchical}, HDT \cite{qi2016hedged} and MDNet \cite{nam2015learning}. To train the network, we use sequences in VOT2013 \cite{kristan2013visual}, VOT2014 \cite{kristan2014visual} and VOT2015 \cite{kristan2015visual}, excluding the videos in OTB \cite{wu2015object}.

Figure~\ref{fig8} illustrates the comparisons of our algorithm with other methods in terms of precision and success plots, respectively. From Figure \ref{fig8}, we can see that our approach outperforms other state-of-the-art trackers in both measures. Besides, to facilitate more detailed analysis, we also report the performance of four lead tracker on different attributes in Figure \ref{fig9}. Experimental results demonstrate that our method can well deal with various challenging factors and consistently outperform the other three trackers in most attributes.

\subsection{Evaluation on VOT2015}
VOT2015 \cite{kristan2015visual} contains 60 image sequences with various challenges. According to VOT challenge protocol in \cite{kristan2015visual},  a tracker is re-initialized whenever failure happens. Two metrics, accuracy and robustness, are utilized to evaluate the performance of trackers. Besides, the VOT challenge also adopts the expected average overlap as a new  evaluation metric, which estimates how accurate the estimated bounding box is after a certain number of frames are processed since initialization. We compare our method with eight state-of-the-art trackers, including DSST \cite{danelljan2014accurate}, DeepSRDCF \cite{danelljan2015learning}, MDNet \cite{nam2015learning}, TGPR \cite{gao2014transfer}, MEEM \cite{zhang2014meem}, MUSTer \cite{hong2015multi}, SAMF \cite{li2014scale}, and LGT \cite{cehovin2013robust}. Our network is pre-trained using sequences from OTB100 \cite{wu2015object}, excluding the sequences in VOT2015 \cite{kristan2015visual} dataset.

\renewcommand\arraystretch{1.1}
\begin{table}[!t]\footnotesize
  \centering
  \caption{The average scores and ranks of accuracy and robustness of different methods on VOT2015 \cite{kristan2015visual}. The top three scores are highlighted in \textcolor{red}{red}, \textcolor{blue}{blue} and \textcolor{green}{green}, respectively.
}
    \begin{tabular}{|c||c|c|c|c|c|}
    \hline
    \multirow{2}[4]{*}{Trackers} & \multicolumn{2}{c|}{Accuracy} & \multicolumn{2}{c|}{Robustness} & Expected \\
    \cline{2-5} & Rank  & Score & Rank  & Score & overlap ratio \\
    \hline
    DSST  & 2.92  & 0.54  & 5.65  & 2.56  & 0.1719 \\
    DeepSRDCF & \textcolor{green}{2.03}  & \textcolor{green}{0.57}  & \textcolor{green}{2.32}  & \textcolor{blue}{1.05}  & \textcolor{green}{0.3181} \\
    LGT   & 5.75  & 0.42  & 4.72  & 2.21  & 0.1737 \\
    MEEM  & 3     & 0.5   & 4.32  & \textcolor{green}{1.85}  & 0.2212 \\
    MUSTer & 2.87  & 0.52  & 4.48  & 2     & 0.1950 \\
    SAMF  & 2.68  & 0.53  & 4.18  & 1.94  & 0.2021 \\
    TGPR  & 3.48  & 0.48  & 5.08  & 2.31  & 0.1938 \\
    MDNet & \textcolor{blue}{1.2}   & \textcolor{blue}{0.6}   & \textcolor{blue}{1.62}  & \textcolor{red}{0.69}  & \textcolor{blue}{0.3783} \\
    \hline
    \hline
    SANet  & \textcolor{red}{1.17}  & \textcolor{red}{0.61}  & \textcolor{red}{1.58}  & \textcolor{red}{0.69}  & \textcolor{red}{0.3895} \\
    \hline
    \end{tabular}%
  \label{tab:addlabel}%
\end{table}%

\begin{figure}[!t]
\centering
\includegraphics[width=72mm,height=45.5mm]{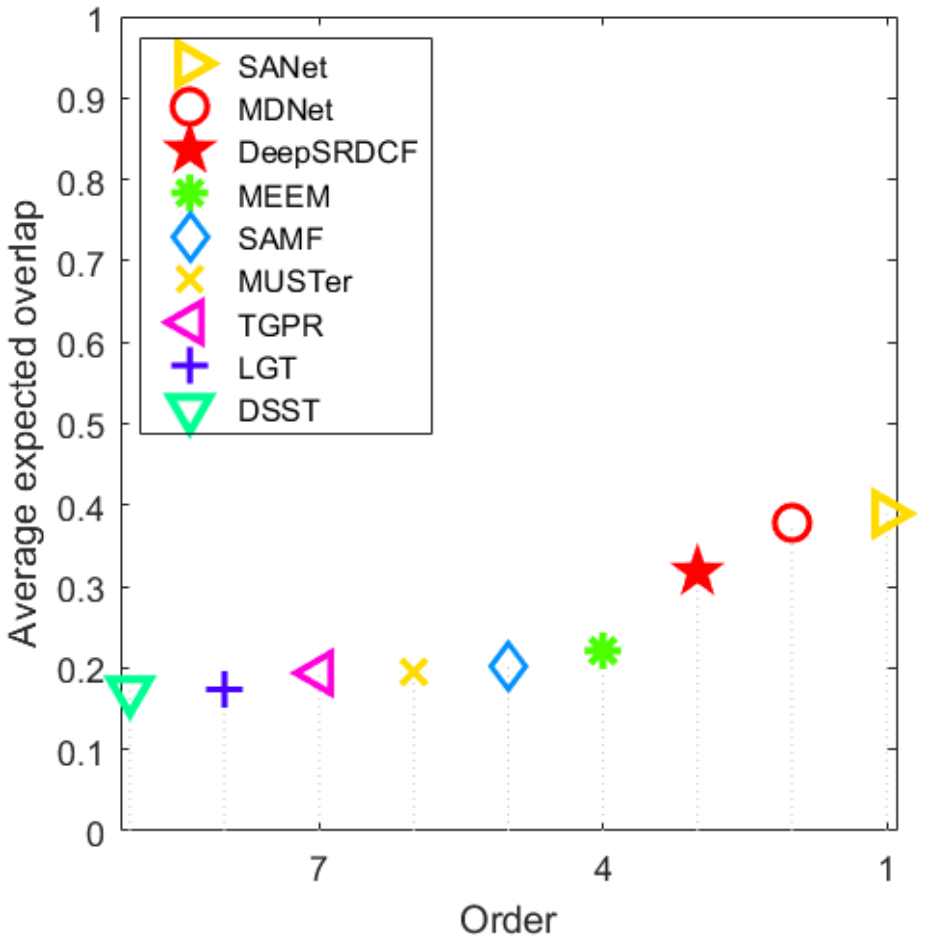}\\
\caption{Expected average overlap ratio graph with trackers ranked from right to left.}\label{fig11}
\end{figure}

Table \ref{tab:addlabel} summarizes the comparison of our tracker with other approaches. From the table we can see that the proposed method outperforms other trackers in all evaluation metrics. Especially, compared with MDNet \cite{nam2015learning}, our tracker demonstrates advances in both accuracy and robustness, showing again the benefits of taking structure information into account. Figure \ref{fig11} visualizes the ranks of trackers on VOT2015 \cite{kristan2015visual} in term of expected overlap ratio.

\section{Conclusion}
\label{secCon}

We present a novel network architecture named SANet for visual tracking by taking into consideration self-structure information of a target object. Different from previous CNNs-based tracking methods, which mainly concentrate on inter-class classification and thus are prone to cause drift in presence of similar distractors, our SANet leverages RNNs to model the structure of target object and combines such structural information with CNNs to learn a discriminative appearance model, which is effective for distinguishing not only background objects of inter-class but also similar distractors of intra-class. Experimental results on three large-scale tracking benchmarks demonstrate the effectiveness of our method.

{\small
\bibliographystyle{ieee}
\bibliography{egbib}
}

\end{document}